\documentclass[10pt,twocolumn,letterpaper]{article}

\PassOptionsToPackage{table}{xcolor}
\usepackage[pagenumbers]{wacv} 

\usepackage{verbatim}
\definecolor{wacvblue}{rgb}{0.21,0.49,0.74}
\usepackage[pagebackref,breaklinks,colorlinks,allcolors=wacvblue]{hyperref}

\usepackage{booktabs}
\usepackage{hyperref}
\usepackage[capitalize]{cleveref}
\usepackage{booktabs}
\usepackage{multirow}
\usepackage{float}

\def\wacvPaperID{2325} 
\def\confName{WACV}
\def\confYear{2026}

\title{\textit{Spectra}: Multi-Scale Implicit Decomposition for Deepfake Detection}

\author{
Dhananjaya Jayasundara\thanks{All authors are with Johns Hopkins University, Corresponding Author (vjayasu1@jh.edu)}
\and
Kartik Narayan\footnotemark[1]
\and
Vishal M. Patel\footnotemark[1]
}

\begin{document}
\maketitle
\begin{abstract}
Deepfakes increasingly undermine trust in digital media, yet existing detectors often struggle to generalize across unseen manipulation techniques, compression levels, and dataset shifts. Most approaches rely on RGB appearance, frequency-domain artifacts, or their combination, learning cues present in the final synthesized image. While effective in-distribution, such cues are frequently generator-specific and can diminish as synthesis models evolve. We argue that a complementary perspective can be obtained through implicit neural representations (INRs), which expose how an image is reconstructed rather than merely how it appears. By fitting an INR to an image and analyzing its internal activations, one can reveal reconstruction dynamics that are difficult to observe in the image domain alone and may capture more intrinsic differences between genuine and synthetic content. To this end, we propose \textit{\textbf{\textit{Spectra}}}, a deepfake detection framework that combines semantic visual representations with artifact-aware cues derived from INR reconstruction. \textit{Spectra} extracts multi-level reconstruction heatmaps from image-specific INRs and jointly models them with semantic features through a bidirectional fusion mechanism. This enables direct interaction between image semantics and reconstruction artifacts, allowing the model to identify forgery cues that are less dependent on specific generation pipelines. Extensive experiments on multiple deepfake benchmarks demonstrate improved robustness and cross-dataset generalization, consistently outperforming detectors based solely on RGB appearance or semantic representations. Furthermore, we show that the proposed representation extends beyond deepfake detection to face presentation attack detection, highlighting its potential as a unified framework for facial attack detection.
\end{abstract}
    
\section{Introduction}

Recent advances in generative modeling have enabled the creation of highly realistic synthetic facial images and videos, commonly referred to as deepfakes. Modern generative architectures, including Generative Adversarial Networks (GANs) \cite{singh2020using,alqahtani2021applications} and diffusion models \cite{croitoru2023diffusion}, can produce photorealistic content that is increasingly difficult for humans to distinguish from authentic media. While these technologies have enabled numerous beneficial applications in content creation and entertainment, they have also raised serious concerns regarding misinformation, identity fraud, impersonation, and media authenticity \cite{hancock2021social,de2023unethical,gilbert2024digital}. As a result, reliable deepfake detection has become an important and active area of research \cite{heidari2024deepfake, altaei2023detection, raza2022novel, frank2020leveraging}.

\begin{figure}[t]
\centering
\includegraphics[width=\columnwidth]{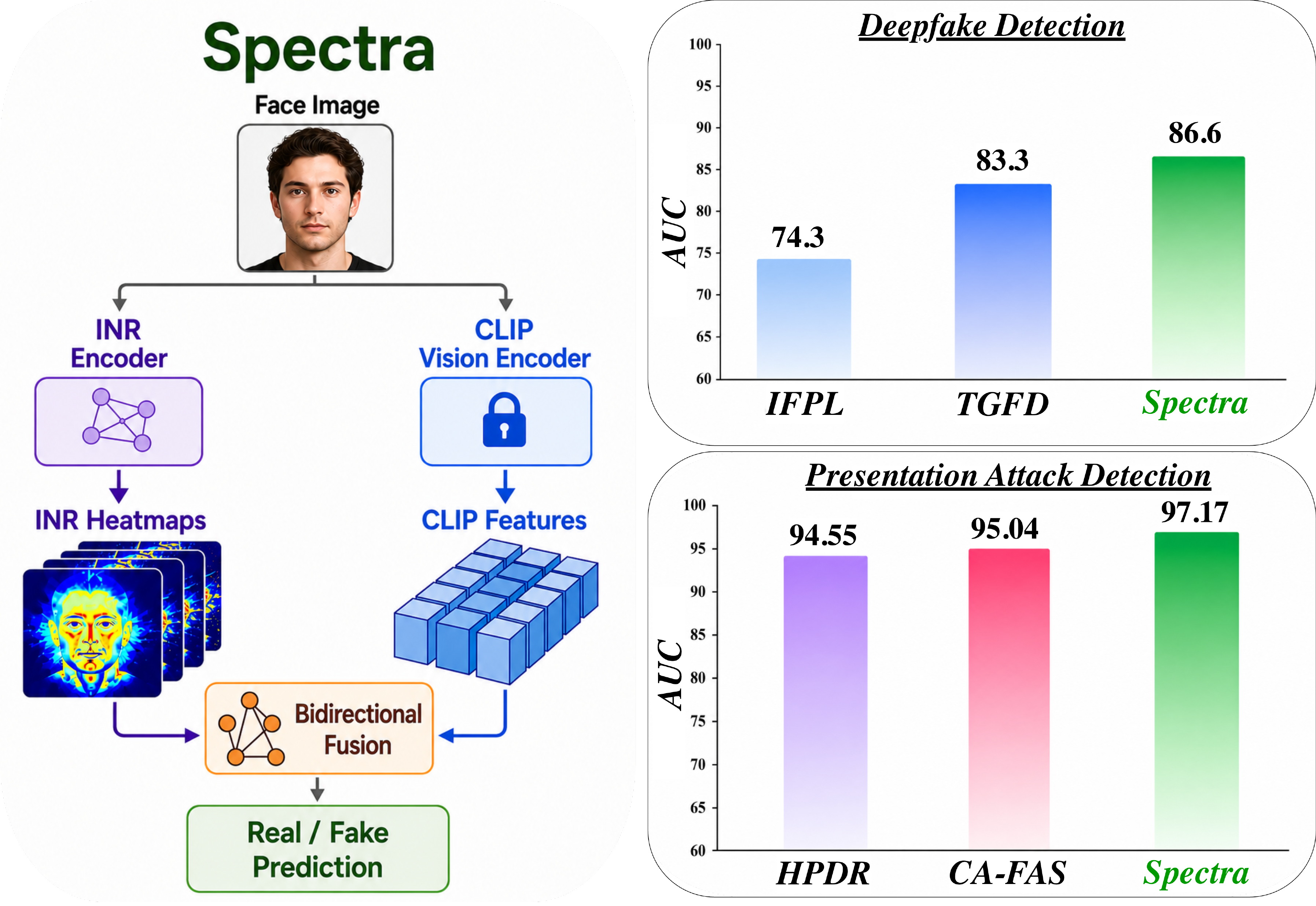}
\vskip-10pt\caption{\textbf{\textit{Spectra} provides a unified framework for two authenticity tasks:}
Spectra employs a unified INR-guided artifact representation to address both deepfake detection and presentation attack detection (PAD). The framework achieves state-of-the-art average AUC on deepfake detection ($86.6\%$) and PAD ($97.17\%$), demonstrating strong generalization across both digital and physical forgery domains.}
\label{fig:unified_results}
\end{figure}

Despite significant progress, a fundamental challenge remains unresolved: \emph{generalization}. Modern deepfake detectors often achieve impressive performance on benchmarks derived from known manipulation pipelines, yet their accuracy can deteriorate substantially when confronted with unseen generators, compression levels, post-processing operations, or dataset shifts. This limitation arises because many detectors learn correlations tied to specific synthesis artifacts rather than properties that are intrinsic to synthetic content itself. As generative models continue to improve and suppress visible artifacts, detectors that rely on such cues become increasingly fragile.

Recent work has demonstrated the effectiveness of large-scale vision-language models such as CLIP \cite{radford2021learning} for deepfake detection \cite{baru2024harnessing,ojha2023towards,chen2025deepfakeclip,yang2026towards}. Owing to large-scale pretraining, CLIP provides rich semantic representations that transfer more effectively across datasets than conventional convolutional features. However, CLIP is fundamentally optimized for semantic understanding rather than forensic analysis. Consequently, it primarily captures \emph{what} an image depicts, while often overlooking subtle traces introduced by the image generation process. As synthetic media becomes increasingly photorealistic, these forensic cues are precisely the signals that become most important for reliable detection.

We argue that a complementary perspective can be obtained through \emph{Implicit Neural Representations (INRs)} \cite{sitzmann2020implicit, essakine2024we, jayasundara2026implicit, jayasundara2025mire, jayasundara2025sinr}. Unlike conventional image representations that operate directly on pixel intensities or their frequency decompositions, INRs model an image as a continuous coordinate-based function and reconstruct it through a hierarchy of nonlinear transformations \cite{essakine2024we, jayasundara2026implicit}. This reconstruction process exposes rich spatial and frequency dynamics that remain hidden in the image domain \cite{yuce2022structured}. Importantly, while two images may appear visually similar, the trajectories followed by an INR during reconstruction can differ substantially. We hypothesize that these reconstruction dynamics encode intrinsic differences between authentic and manipulated content, providing a source of forensic evidence that is less dependent on the particular synthesis pipeline. To explore this hypothesis, we extract multi-level heatmaps from trained INRs and use them to characterize how image information is represented across reconstruction stages. These heatmaps reveal localized structural and frequency-sensitive responses that are complementary to semantic visual features and offer a novel artifact-aware representation for deepfake detection.

Building upon this insight, we propose \textbf{\textit{Spectra}}, a deepfake detection framework that unifies semantic understanding with reconstruction-based forensic analysis. \textit{Spectra} leverages a frozen CLIP vision encoder to capture high-level semantic content while simultaneously extracting multi-level heatmaps from image-specific INRs that expose reconstruction dynamics hidden from the image domain. These heatmaps are processed through complementary spatial and frequency encoders, enabling the model to capture both localized structural irregularities and frequency-sensitive inconsistencies that emerge during reconstruction. To bridge the gap between semantics and artifacts, \textit{Spectra} employs bidirectional attention mechanisms that allow semantic representations to focus on manipulation-sensitive regions while enabling reconstruction cues to selectively retrieve relevant semantic context (See \Cref{fig:unified_results}). Rather than treating artifacts and semantics as independent signals, \textit{Spectra} explicitly models their interactions, producing a unified representation that captures both \emph{what} an image depicts and \emph{how} it is reconstructed. We argue that this combination yields a more fundamental characterization of synthetic content, reducing reliance on generator-specific artifacts and improving robustness to unseen manipulations.

Extensive experiments across multiple deepfake benchmarks demonstrate that \textit{Spectra} consistently outperforms strong semantic-only baselines and achieves superior cross-dataset generalization. Beyond performance gains, our analysis reveals that INR-derived reconstruction representations provide complementary forensic evidence that is largely absent from conventional image-domain features. These findings establish reconstruction dynamics as a promising new direction for robust and generalizable media forensics.

\begin{figure*}[t]
    \centering
    \includegraphics[width=.85\textwidth]{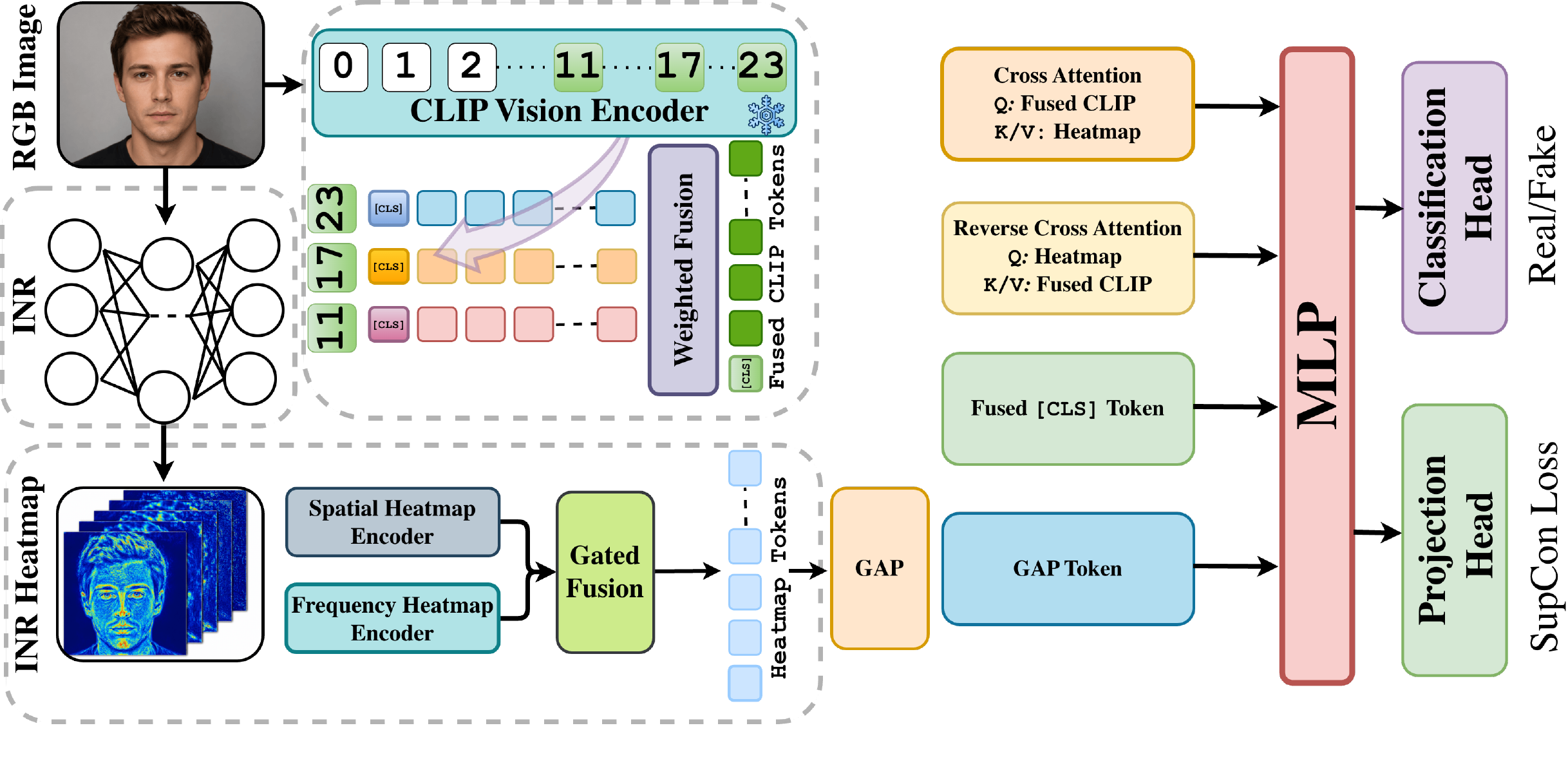}
   \vskip-20pt \caption{\textbf{Overview of \textit{\textit{Spectra}}.} Given an input face image, a frozen CLIP vision encoder extracts multiscale semantic representations from selected transformer layers, while an INR fitted to the same image produces layer-wise heatmaps that capture reconstruction dynamics and frequency-specific artifacts. The heatmaps are processed by spatial and frequency encoders and fused through a gated aggregation module to obtain artifact-aware heatmap tokens. In parallel, multiscale CLIP features are combined through weighted fusion to produce semantic tokens. To model interactions between semantic content and reconstruction artifacts, we employ both cross-attention (CLIP queries heatmap features) and reverse cross-attention (heatmap queries CLIP features). The resulting representations, together with global CLIP and heatmap descriptors, are fused and passed to an MLP for real/fake classification. A projection head attached to the shared representation enables supervised contrastive learning, encouraging stronger separation between genuine and manipulated samples.}
\label{fig:main_img}
    \label{fig:main_img}
\end{figure*}

\section{Method}


An overview of the proposed \textit{\textit{Spectra}} framework is shown in \Cref{fig:main_img}. Given an input image $I$, we first obtain two complementary representations. A pretrained CLIP vision encoder extracts multi-scale semantic features from selected transformer layers, while an image-specific INR produces a multi-level decomposition that captures reconstruction behavior across different representation stages (See \Cref{fig:discriminative}). The resulting decomposition maps are processed through spatial and frequency branches to generate artifact-aware heatmap representations.

The semantic and artifact streams are subsequently fused through a bidirectional artifact-semantic interaction module. Specifically, semantic features attend to reconstruction-aware artifact representations, while artifact features simultaneously retrieve complementary semantic context. This mutual interaction produces a unified representation that combines high-level image understanding with reconstruction-aware cues. Finally, the fused features are used for both binary classification and supervised contrastive learning, enabling the network to learn discriminative embeddings for robust facial attack detection.

\subsection{Dataset Preparation}
\label{sec:dataset}

To build a robust dataset for training and evaluation, we follow a systematic preprocessing pipeline consisting of frame extraction, face detection, and face alignment. We first extract frames from each video and subsequently detect faces using the RetinaFace~\cite{deng2020retinaface} detector. The detected faces are cropped according to the predicted bounding boxes and aligned using five facial landmark keypoints. Alignment is performed through a warp-affine transformation to standardize facial geometry across samples. Following alignment, all face crops are resized to a fixed resolution. \textit{\textit{Spectra}} is trained on the \(c_{23}\) version of the FaceForensics++ (FF++) dataset~\cite{rossler2019faceforensics++}, which contains realistic video compression artifacts. For training, we sample 10 frames from each fake video and 40 frames from each real video. Since a primary objective of deepfake detection is robust generalization to unseen forgery methods, we evaluate the trained model on four out-of-distribution (OOD) benchmarks: \textbf{Celeb-DF v1} (CDF$_{\mathrm{v1}}$)~\cite{li2020celeb}, \textbf{Celeb-DF v2} (CDF$_{\mathrm{v2}}$)~\cite{li2019celeb}, \textbf{FaceShifter} (FSh)~\cite{rossler2019faceforensics++}, and \textbf{Deep Fake Detection} (DFD)~\cite{rossler2019faceforensics++}.

\subsection{Multi-Level Implicit Decomposition}

Most existing deepfake detectors operate directly on RGB images, learning discriminative representations solely from pixel-space visual information. While effective, RGB representations expose only the final image appearance and reveal nothing about how that appearance was internally constructed. We instead seek a reconstruction-aware representation that surfaces the internal dynamics of image formation, which we hypothesize differ systematically between authentic and manipulated images.

Given an image $I \in \mathbb{R}^{H \times W \times 3}$, each pixel is a coordinate--color pair $(\mathbf{x}_i, \mathbf{c}_i)$, with $\mathbf{x}_i = (u_i, v_i) \in \mathbb{R}^2$ the spatial coordinate and $\mathbf{c}_i = (r_i, g_i, b_i) \in \mathbb{R}^3$ the corresponding RGB value. An INR parameterizes the image as a continuous function $f_\theta: \mathbb{R}^2 \rightarrow \mathbb{R}^3$ mapping coordinates to color, $f_\theta(\mathbf{x}_i) \approx \mathbf{c}_i$, with $\theta$ fit by minimizing the reconstruction objective $\mathcal{L}_{\text{INR}} = \sum_i \| f_\theta(\mathbf{x}_i) - \mathbf{c}_i \|_2^2$. For \textit{\textit{Spectra}}, $f_\theta$ is a SIREN \cite{sitzmann2020implicit}, whose periodic activations make it especially well suited to representing natural image signals. 

Unlike conventional image encoders, an INR reconstructs an image through a hierarchy of latent activations. Consistent with the \textit{Spectra}l bias of neural networks \cite{rahaman2019spectral}, this reconstruction progresses from coarse image structure toward increasingly fine detail across hidden layers. To expose these intermediate representations, we extract activations from each hidden SIREN layer and project them into image-space heatmaps via principal component analysis (PCA), isolating, at each layer, the dominant direction of image-specific variation among the activations.

The resulting decomposition is visualized in \Cref{fig:discriminative}, contrasting a genuine image (top row) with a deepfake (bottom row) from CDF\textsubscript{v2}. Even though both are reconstructed faithfully by the SIREN (rightmost column), their internal representations evolve differently across layers. At the first decomposition level, both examples exhibit structured, wave-like patterns characteristic of the SIREN's sinusoidal activations; here the two are visually similar, reflecting primarily low-level, coordinate-dependent responses. As reconstruction progresses, the representations diverge: at the second level, the genuine image develops coherent facial structure, with smooth responses organizing around semantically meaningful regions such as the eyes, nose, and mouth, while the manipulated image exhibits less organized activations and stronger localized variation. This divergence sharpens at the deepest level we extract, where the genuine image retains coherent spatial organization while the deepfake displays fragmented responses, distorted contours, and amplified edge-like structures. Notably, many of these differences are difficult to observe directly in the RGB image or its frequency based representation. Even when both images yield visually similar reconstructions, their intermediate activations reveal substantially different representation pathways, evidence that reconstruction dynamics carry discriminative information largely absent from the final reconstructed image. See supplementary for more reconstruted results. 

Motivated by this, we extract PCA-based heatmaps from the first $K$ hidden SIREN layers, $\mathcal{H} = \{H_1, H_2, \ldots, H_K\}$, where each $H_k$ captures the representation state at one stage of reconstruction. The divergence in \Cref{fig:discriminative} further suggests that manipulations surface not only within individual levels but in how representations \emph{evolve} between them; we capture this with inter-level difference maps $D_k = |H_{k+1} - H_k|$, which highlight regions of significant representational change between consecutive stages and encode dynamics invisible to any single level in isolation. The full multi-level decomposition is therefore $\mathcal{S} = \{H_1, \ldots, H_K, D_1, \ldots, D_{K-1}\}$; for the three-level decomposition ($K{=}3$) used in our experiments, this yields five channels, $\mathcal{S} = \{H_1, H_2, H_3, D_1, D_2\}$. Together, the heatmaps and difference maps form a compact reconstruction-aware representation that captures both the state of reconstruction and its evolution across levels, exposing structural and frequency-sensitive inconsistencies introduced during image synthesis. While per-image INR optimization has been shown to produce strong reconstruction quality \cite{erkocc2023hyperdiffusion,de2023deep}, it requires iterative optimization and can be computationally costly. To improve efficiency, we instead employ a hypernetwork-based approach that directly predicts the parameters of a SIREN INR, eliminating the need for per-image optimization. Further details are provided in the supplementary material.

\begin{figure}[t]
    \centering
    \includegraphics[width=.9\columnwidth]{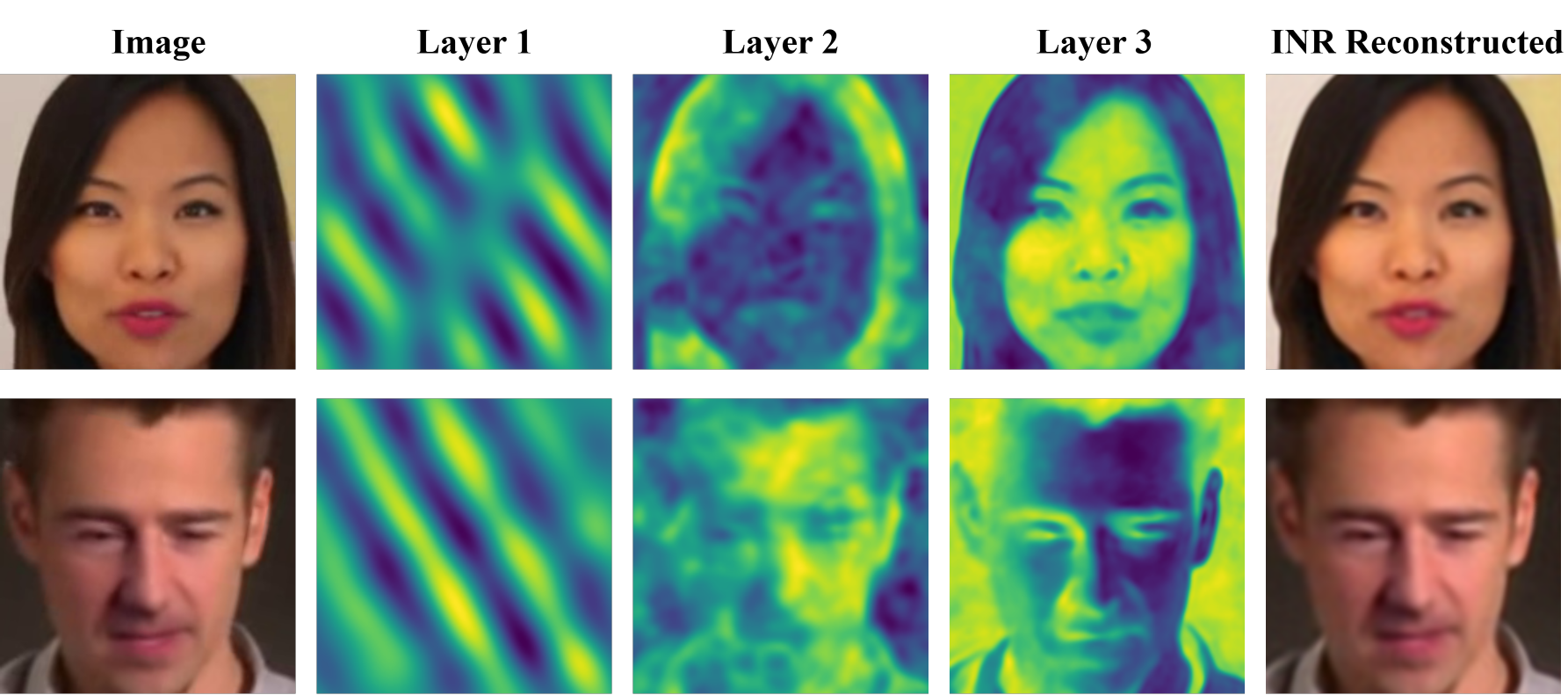}
    \vskip-5pt\caption{\textbf{Layer-wise PCA decomposition of SIREN reconstruction dynamics.} Top: a genuine face from CDF\textsubscript{v2}; bottom: a deepfake. Although both are reconstructed with comparable fidelity (rightmost column), their internal activations diverge sharply across layers}
\label{fig:discriminative}
    \label{fig:discriminative}
\end{figure}

\subsection{Reconstruction-Aware Representation \\Learning}
\label{sec:reconstruction_aware}

The multi-level implicit decomposition encodes complementary spatial and spectral evidence about how an image is reconstructed by the SIREN. Local inconsistencies may surface as irregular facial structure, blending boundaries, texture discontinuities, or unstable responses around high-detail regions, while other manipulation traces are more apparent in the spectral behavior of the decomposition maps. Relying on either cue alone therefore risks discarding useful forensic evidence.

Let $\mathcal{S}\in\mathbb{R}^{C\times H\times W}$ denote the multi-level implicit decomposition, where $C$ is the number of decomposition channels. To capture localized reconstruction patterns, we process $\mathcal{S}$ using a spatial encoder $E_s(\cdot)$ and obtain $F_s=E_s(\mathcal{S})$, where $F_s\in\mathbb{R}^{d\times h\times w}$. In parallel, we compute the log-magnitude spectrum of the decomposition maps, defined as $\mathcal{F}=\log(1+|\mathrm{FFT}(\mathcal{S})|)$, and process it using a frequency encoder $E_f(\cdot)$ to obtain $F_f=E_f(\mathcal{F})$, where $F_f\in\mathbb{R}^{d\times h\times w}$. The spatial branch emphasizes localized structural responses in the INR decomposition, whereas the frequency branch captures spectral irregularities in the reconstruction behavior.

As the relative importance of spatial and frequency cues can vary across images and manipulation types, we fuse them adaptively rather than using a fixed combination. Global descriptors are first extracted from $F_s$ and $F_f$ using global average pooling. A lightweight gating network then predicts a channel-wise weighting vector $g=\sigma(W_g[\mathrm{GAP}(F_s),\mathrm{GAP}(F_f)])$, where $g\in\mathbb{R}^{d}$, $\sigma(\cdot)$ denotes the sigmoid function, and $[\cdot,\cdot]$ denotes concatenation. The final reconstruction-aware representation is computed as $F_h=g\odot F_s+(1-g)\odot F_f$, where $\odot$ denotes element-wise multiplication. This adaptive fusion allows \textit{\textit{Spectra}} to emphasize the most informative spatial or spectral reconstruction cues for each input image.

\subsection{Multi-Scale Semantic Representation Learning}

\label{sec:multiscale_semantic}

While the proposed multi-level decomposition provides reconstruction-aware forensic cues, it contains limited semantic understanding of the underlying image content. In particular, reconstruction irregularities are often most informative when interpreted in the context of facial structure, identity, and appearance. To provide this complementary information, we leverage semantic representations extracted from a pretrained CLIP vision transformer.

A key characteristic of transformer-based vision models is that semantic information emerges progressively across network depth. Earlier layers primarily encode local visual patterns and texture statistics, intermediate layers capture object parts and structural relationships, while deeper layers produce highly abstract semantic representations. Consequently, relying solely on the final transformer layer may discard useful information present at intermediate representation levels. This is particularly relevant for facial forensics, where manipulation traces can manifest at multiple semantic scales, ranging from localized texture inconsistencies to higher-level identity and facial structure deviations.

Given an input image $I$, let $L = \{l_1, \ldots, l_S\}$ denote the set of selected transformer layers. We select $L$ to span early, intermediate, and late representation depths rather than relying on the final layer alone, for a 24-layer CLIP ViT-L/14, this corresponds to layers $\{11, 17, 23\}$, sampled at approximately half-depth, three-quarter-depth, and the final layer respectively. We find this combination to outperform any individual layer or partial subset (See \Cref{tab:ablation-multiscale-blocks}), consistent with the intuition that forgery-relevant cues can emerge at multiple semantic depths. For each layer $l \in L$, the CLIP encoder produces a token sequence $Z^{(l)} = \{z_{\mathrm{cls}}^{(l)}, z_1^{(l)}, \ldots, z_N^{(l)}\} \in \mathbb{R}^{(N+1) \times d_l}$, where $z_{\mathrm{cls}}^{(l)}$ denotes the class token, $\{z_i^{(l)}\}_{i=1}^{N}$ are patch tokens, $N$ is the number of image patches, and $d_l$ is the embedding dimension at layer $l$. Since features extracted from different transformer depths may exhibit distinct distributions, each representation is projected into a common embedding space through a learnable projection function $P_l: \mathbb{R}^{d_l} \rightarrow \mathbb{R}^{d}$.

Rather than treating all semantic scales equally, we learn their relative importance through adaptive scale fusion. Specifically, the projected representations are combined as $Z_s = \sum_{l \in L} \alpha_l P_l(Z^{(l)})$, where $\alpha_l = \exp(\beta_l) / \sum_{j \in L} \exp(\beta_j)$ is a learnable normalized weight associated with layer $l$, satisfying $\sum_{l \in L} \alpha_l = 1$. This formulation allows the network to automatically emphasize the semantic scales that are most informative for forgery detection.

The resulting representation $Z_s = \{z_{\mathrm{cls}}, z_1, \ldots, z_N\} \in \mathbb{R}^{(N+1) \times d}$ consists of a global semantic token $z_{\mathrm{cls}}$ and a set of semantic patch tokens $\{z_i\}_{i=1}^{N}$. The global token provides an image-level semantic summary, while the patch tokens preserve localized semantic information. Together, these representations provide a semantically rich description of the input image and serve as the semantic stream used by the subsequent artifact-semantic fusion module.

\subsection{Bidirectional Artifact-Semantic Fusion}

The reconstruction-aware representation and the semantic representation provide complementary evidence for forgery detection. The former captures how an image is reconstructed and highlights manipulation-sensitive response patterns, while the latter provides contextual understanding of facial structure, identity, and appearance. Neither may be sufficient on its own, reconstruction cues lack semantic context, making it difficult to determine whether an irregular response corresponds to a meaningful facial region or an incidental pattern, while semantic features are optimized for visual understanding and may overlook subtle manipulation traces. We therefore seek a representation that jointly reasons over both modalities.

To this end, we introduce a bidirectional artifact-semantic fusion mechanism. Rather than concatenating semantic and reconstruction-aware features outright, the module explicitly models their interaction from two complementary directions, semantic features retrieve relevant reconstruction evidence from the artifact representation, and reconstruction-aware features in turn guide the aggregation of semantic information toward manipulation-sensitive regions. This bidirectional exchange lets the model capture both \emph{what} an image depicts and \emph{how} it was reconstructed.

Let $F_h \in \mathbb{R}^{C_h \times H_h \times W_h}$ denote the reconstruction-aware feature map from \cref{sec:reconstruction_aware} (where $C_h$ is the channel count and $(H_h, W_h)$ the spatial resolution), reshaped into a sequence of $M = H_h W_h$ reconstruction-aware tokens $T_h = \{t_1, \ldots, t_M\} \in \mathbb{R}^{M \times C_h}$. Recall from \cref{sec:multiscale_semantic} the multi-scale semantic representation $Z_s = \{z_{\mathrm{cls}}, z_1, \ldots, z_N\}$, with class token $z_{\mathrm{cls}} \in \mathbb{R}^{C_s}$ and patch tokens $T_s = \{z_1, \ldots, z_N\} \in \mathbb{R}^{N \times C_s}$. Both representations are projected into a shared attention space of dimension $C_a$, distinct from the semantic embedding dimension $C_s = d$ and the reconstruction-aware channel count $C_h$, for interaction.\\
\noindent{\bf{Reconstruction-aware semantic aggregation}}
The first interaction stage lets semantic features retrieve reconstruction-aware evidence from the artifact representation. A learnable aggregation token $a \in \mathbb{R}^{C_a}$ is prepended to the projected semantic patch tokens, forming the query sequence $Q = [a;\, W_Q T_s] \in \mathbb{R}^{(N+1) \times C_a}$. The reconstruction-aware tokens are projected into keys and values, $K = W_K T_h$ and $V = W_V T_h \in \mathbb{R}^{M \times C_a}$, and cross-attention is computed as $A = \mathrm{Softmax}\big(QK^\top / \sqrt{C_a}\big)$, $O = AV \in \mathbb{R}^{(N+1) \times C_a}$. The aggregation token attends jointly with the semantic patch tokens but, unlike conventional global pooling, learns to selectively retrieve whichever reconstruction-aware evidence is most relevant to the image's semantic content. We take the final output $O_0$ and refine it with a residual feed-forward block, $f_{\mathrm{attn}} = O_0 + \mathrm{FFN}(O_0) \in \mathbb{R}^{C_a}$, yielding a compact semantic descriptor enriched with reconstruction-aware forensic evidence.\\
\noindent{\bf{Reconstruction-guided semantic aggregation.}}
The converse interaction is equally important, reconstruction irregularities often indicate which semantic regions deserve closer attention. Rather than treating all semantic patches equally, we use the reconstruction-aware tokens to guide semantic aggregation directly. For each reconstruction-aware token $t_i \in T_h$, a learnable scoring function predicts an importance score $\alpha_i = \sigma(W_w t_i + b_w)$, with $W_w \in \mathbb{R}^{1 \times C_h}$ and bias $b_w$. Since the number of reconstruction-aware and semantic tokens generally differs ($M \neq N$), we linearly interpolate the resulting score sequence to a semantic guidance vector $\tilde{\alpha} = \{\tilde{\alpha}_1, \ldots, \tilde{\alpha}_N\}$ matching the semantic patch count, then aggregate the semantic tokens accordingly: $z_{\mathrm{guide}} = \sum_{i=1}^{N} \tilde{\alpha}_i \, \mathrm{LN}(z_i)$, where $\mathrm{LN}(\cdot)$ denotes layer normalization \cite{ba2016layer}. A learnable projection maps this aggregate into the shared attention space, $f_{\mathrm{guide}} = W_g z_{\mathrm{guide}} + b_g \in \mathbb{R}^{C_a}$. The resulting descriptor up-weights semantic regions that coincide with strong reconstruction responses, so facial structures exhibiting anomalous reconstruction behavior contribute more strongly to the final representation letting the model focus on regions that are simultaneously semantically meaningful and manipulation-sensitive.\\
\noindent{\bf{Unified representation.}}
The final image representation combines complementary global and interaction-aware descriptors: the global semantic token $z_{\mathrm{cls}}$, the artifact-aware semantic descriptor $f_{\mathrm{attn}}$, the artifact-guided semantic descriptor $f_{\mathrm{guide}}$, and a global reconstruction-aware descriptor $f_{\mathrm{global}} = \mathrm{GAP}(F_h)$. These are concatenated into the fused representation $f_{\mathrm{fuse}} = [z_{\mathrm{cls}}, f_{\mathrm{attn}}, f_{\mathrm{guide}}, f_{\mathrm{global}}]$, jointly encoding image semantics, reconstruction-aware evidence, and the bidirectional interaction between them.

\subsection{Joint Representation Learning}

Rather than relying on any single descriptor, \textit{\textit{Spectra}} integrates the four complementary views in $f_{\mathrm{fuse}}$ to jointly reason about image semantics and manipulation artifacts. This representation is passed through a bottleneck projection network $\phi(\cdot)$ (MLP), yielding a compact embedding $e = \phi(f_{\mathrm{fuse}})$ that serves as the final representation for downstream prediction. Because $e$ jointly encodes semantic content and reconstruction-aware artifact cues rather than either in isolation, the same representation transfers naturally across forensic tasks, we use it for both deepfake detection and presentation attack detection, with only the training distribution changing between the two.

\subsection{Training Objective}

The final embedding $e$ is used for binary classification through a head producing class probabilities $p = \mathrm{Softmax}(W_c e + b_c)$, supervised with cross-entropy loss $\mathcal{L}_{\mathrm{CE}}$. To improve robustness to manipulation-specific artifacts, we apply mixup~\cite{zhang2017mixup} to the classification branch: for $\lambda \sim \mathrm{Beta}(\alpha, \alpha)$ and a random batch permutation $\sigma$, every input stream feeding the classifier, semantic tokens, reconstruction-aware tokens, and multi-scale CLIP features, is interpolated with the \emph{same} $(\lambda, \sigma)$ pair, keeping the mixed sample consistent across modalities. The loss is taken against the correspondingly interpolated targets, $\mathcal{L}_{\mathrm{CE}} = \lambda\, \mathrm{CE}(p, y_i) + (1-\lambda)\, \mathrm{CE}(p, y_{\sigma(i)})$. Cross-entropy alone may not constrain the geometry of the feature space, leaving same-class samples potentially dispersed and limiting generalization to unseen manipulations. We therefore add supervised contrastive learning~\cite{khosla2020supervised}, computed on the \emph{unmixed} batch, since mixup's fractional labels are incompatible with discrete class supervision. The embedding is projected to $z = \psi(e)$, and for a batch of $B$ samples with $P(i)$ the indices sharing anchor $i$'s label (excluding $i$) and $A(i) = \{1,\ldots,B\}\setminus\{i\}$ all others, the loss is $\mathcal{L}_{\mathrm{SupCon}} = \sum_{i=1}^{B} \frac{-1}{|P(i)|} \sum_{p \in P(i)} \log \frac{\exp(z_i^{\top} z_p / \tau)}{\sum_{a \in A(i)} \exp(z_i^{\top} z_a / \tau)}$, pulling same-class embeddings together while pushing different classes apart. The full objective is $\mathcal{L} = \mathcal{L}_{\mathrm{CE}} + \lambda_{\mathrm{sc}}\, \mathcal{L}_{\mathrm{SupCon}}$, with mixup regularizing the decision boundary and supervised contrastive learning organizing the embedding space together yielding a feature space robust to manipulation technique, compression, and dataset shift.

\section{Experiments}
\label{sec:experiments}


To evaluate the effectiveness of \textit{\textit{Spectra}}, we conduct experiments across multiple deepfake detection benchmarks. The training set consists of frames sampled from videos generated using four widely used face manipulation techniques: \textbf{Deepfakes}, \textbf{Face2Face}, \textbf{FaceSwap}, and \textbf{NeuralTextures}. These methods represent diverse manipulation strategies and provide a challenging training distribution. The evaluation datasets, described in Section~\ref{sec:dataset}, differ from the training data in terms of manipulation technique, generation pipeline, and visual characteristics, enabling a rigorous assessment of cross-dataset generalization. Following common practice, performance is measured using the Area Under the ROC Curve (AUC).

\Cref{tab:cross_dataset_auc} summarizes the cross-dataset performance of \textit{\textit{Spectra}} against state-of-the-art methods on four widely-used out-of-distribution deepfake detection benchmarks (``--'' denotes results not reported in prior work). \textit{\textit{Spectra}} achieves the best or second-best AUC on every benchmark, and the best average AUC overall, demonstrating consistently strong generalization under distribution shift rather than performance confined to a single dataset or manipulation family. On CDF\textsubscript{v1}, \textit{\textit{Spectra}} attains an AUC of 0.829, outperforming the previous best method, MUF (0.807), by a relative improvement of \textbf{2.73\%}. The gain is more pronounced on CDF\textsubscript{v2}, where \textit{\textit{Spectra}} reaches 0.841 against TGFD's 0.794, which is a \textbf{5.92\%} relative improvement, and most pronounced on FSh, where \textit{\textit{Spectra}}'s 0.858 surpasses the best prior result, WaveletCLIP (0.732), by \textbf{17.21\%}. On DFD, \textit{\textit{Spectra}} is narrowly second-best at 0.896, trailing TGFD's 0.905 by under one AUC point; despite this single exception, \textit{\textit{Spectra}} achieves the highest average AUC (0.856) across all four benchmarks, outperforming the next-best method, TGFD (0.833), by \textbf{2.76\%}. Against CLIP-based methods specifically, \textit{\textit{Spectra}} improves over the strongest baseline, WaveletCLIP (0.749 average), by \textbf{14.29\%}, underscoring that the gain stems not merely from using a CLIP backbone but from how \textit{\textit{Spectra}} couples it with reconstruction-aware artifact evidence. Taken together, these results show that \textit{\textit{Spectra}} delivers consistently strong cross-dataset performance without dataset-specific tuning, establishing a new state of the art on three of the four benchmarks evaluated.

\begin{figure}[t]
    \centering
    \includegraphics[width=\columnwidth]{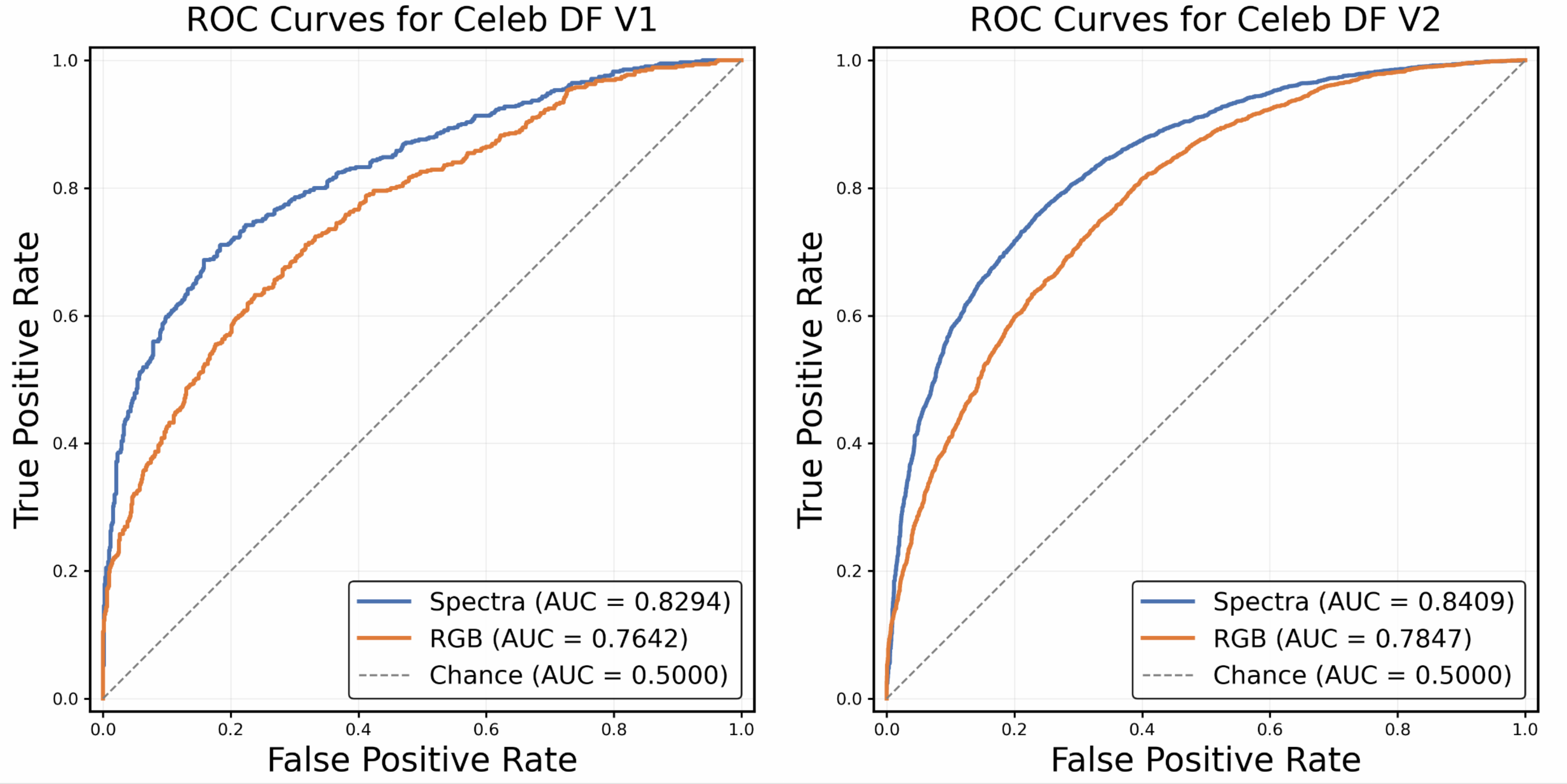}
    \vskip-5pt\caption{\textbf{ROC curves.} Comparison of the proposed \textit{\textit{Spectra}} model and the RGB-only ablation on the Celeb-DF v1, and v2 benchmarks.}
    \label{fig:roc_curve}
\end{figure}

\subsection{How Well Does \textit{\textit{Spectra}} Separate Classes?}

\begin{figure}[h]
    \centering
    \includegraphics[width=\linewidth]{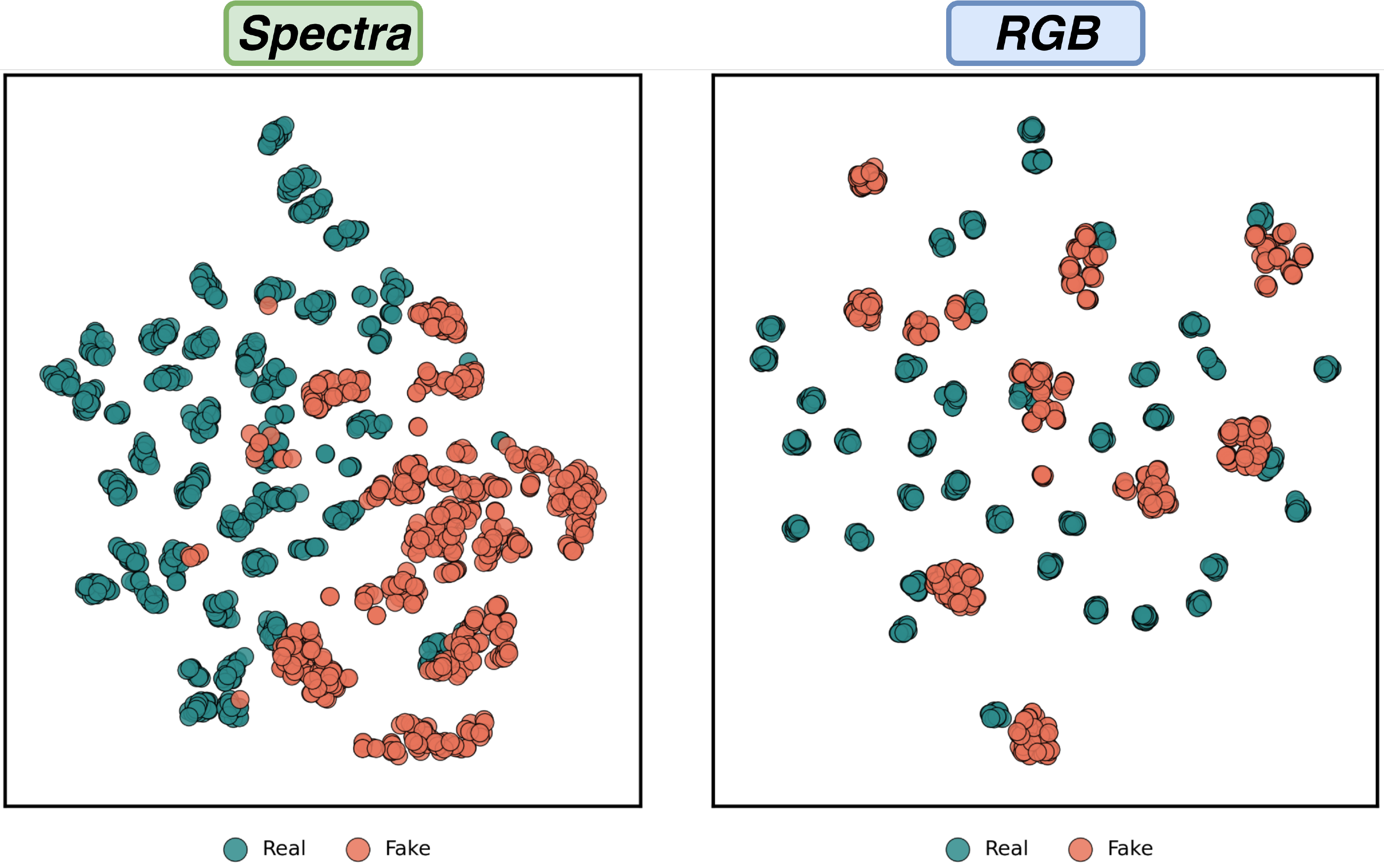}
   \vskip-5pt \caption{\textbf{t-SNE visualization of feature embeddings on the CDFv1 dataset.} The proposed \textit{\textit{Spectra}} model produces more discriminative feature representations than the \textit{RGB}-only baseline, resulting in improved separation between real and fake samples.}
    \label{fig:cdf_v1_tsne}
\end{figure}

Beyond AUC, we examine how well \textit{\textit{Spectra}}'s learned representations separate real and fake samples directly in feature space. To isolate the contribution of the reconstruction-aware heatmap, we compare against an RGB ablation, which is, the identical architecture in \Cref{fig:main_img}, but with the heatmap branch fed the raw RGB image in place of the INR decomposition. We report the Silhouette score (higher is better) \cite{rousseeuw1987silhouettes} and Davies--Bouldin index (lower is better) \cite{davies1979cluster} on the bottleneck embedding of both variants (MLP), and visualize the embedding via t-SNE on CDF\textsubscript{v1} (\Cref{fig:cdf_v1_tsne}). As shown in \Cref{tab:cluster_metrics}, \textit{\textit{Spectra}} improves Silhouette by \textbf{$2.4$--$3.2\times$} and Davies--Bouldin by \textbf{$1.6$--$1.9\times$} over the RGB ablation across all four benchmarks. The t-SNE visualization confirms this, under \textit{\textit{Spectra}}, fake samples form a few large, well-defined clusters while real samples remain more dispersed, whereas the RGB ablation scatters both classes into many small, mixed clusters.
Together, these results show that the reconstruction-aware representation does not merely shift the decision boundary, it reorients the embedding space itself into a more linearly separable configuration, easing the downstream classifier's task. Further, the corresponding ROC curves are shown in \Cref{fig:roc_curve}.

\begin{table}[t]
\centering
\footnotesize
\setlength{\tabcolsep}{3pt}
\caption{\textbf{Cross-dataset frame-level AUC.}
Best and second-best results are shown in
\textbf{bold} and \underline{underlined}, respectively.}
\label{tab:cross_dataset_auc}

\begin{tabular}{lccccc}
\toprule
Method & CDF$_{v1}$ & CDF$_{v2}$ & FSh & DFD & Avg \\
\midrule

\multicolumn{6}{c}{\emph{General Methods}} \\
\midrule

MesoNet'18 \cite{afchar2018mesonet}      & 0.735 & 0.609 & 0.566 & 0.548 & 0.615 \\
MesoInc.'18 \cite{afchar2018mesonet}     & 0.736 & 0.696 & 0.643 & 0.607 & 0.671 \\
Xception'18  \cite{chollet2017xception}    & 0.779 & 0.736 & 0.624 & 0.816 & 0.739 \\

DSP-FWA'18  \cite{li2018exposing}     & 0.789 & 0.668 & 0.555 & 0.740 & 0.688 \\
EffNet'19 \cite{tan2019efficientnet}      & 0.790 & 0.748 & 0.616 & 0.815 & 0.742 \\

Capsule'19 \cite{nguyen2019capsule}    & 0.790 & 0.747 & 0.646 & 0.684 & 0.717 \\

CNN-Aug'20 \cite{wang2020cnn}     & 0.742 & 0.702 & 0.598 & 0.646 & 0.672 \\
Face X-ray'20 \cite{li2020face}  & 0.709 & 0.678 & 0.655 & 0.766 & 0.702 \\
FFD'20 \cite{dang2020detection}         & 0.784 & 0.744 & 0.605 & 0.802 & 0.734 \\
F$^3$-Net'20 \cite{qian2020thinking}   & 0.776 & 0.735 & 0.591 & 0.798 & 0.725 \\
SRM'21 \cite{luo2021generalizing}         & 0.792 & 0.755 & 0.601 & 0.812 & 0.740 \\
CORE'22 \cite{ni2022core}        & 0.779 & 0.743 & 0.603 & 0.802 & 0.732 \\
RECCE'22 \cite{cao2022end}       & 0.767 & 0.731 & 0.609 & 0.812 & 0.730 \\
UCF'23 \cite{yan2023ucf}         & 0.779 & 0.752 & 0.646 & 0.807 & 0.746 \\
F-G'24 \cite{lin2024preserving}         & 0.744 & -- & -- & 0.848 & 0.796 \\

DF-GDD'24~\cite{ye2024decoupling}
                & -- & 0.769 & -- & 0.830 & 0.800 \\
                
TGFD'25~\cite{bai2025towards}
                & 0.799 & \underline{0.794} & -- & \textbf{0.905} & \underline{0.833} \\
CSFI'25~\cite{jing2025constructing}
                & 0.804 & -- & -- & -- & 0.804 \\
WA-ViT CDC'25~\cite{badr2025wavit}
                & 0.786 & -- & -- & -- & 0.786 \\
MUF'26~\cite{li2026multiple}
                & \underline{0.807} & 0.768 & -- & 0.819 & 0.798 \\

\midrule
\multicolumn{6}{c}{\emph{CLIP-Based Methods}} \\
\midrule

CLIP'23\cite{ojha2023towards}        & 0.743 & 0.750 & 0.730 & -- & 0.741 \\
WaveletCLIP'25\cite{baru2024harnessing}  & 0.756 & 0.759 & \underline{0.732} & -- & 0.749 \\
IFPL'26~\cite{yang2026towards}
                & 0.804 & 0.722 & -- & 0.702 & 0.743 \\

\rowcolor{green!10}
\textbf{\textit{Spectra}}
                & \textbf{0.829}
                & \textbf{0.841}
                & \textbf{0.858}
                & \underline{0.896}
                & \textbf{0.856} \\

\bottomrule
\end{tabular}
\end{table}

\begin{table}[t]
\centering
\caption{\textbf{Cluster-separation metrics.} Silhouette ($\uparrow$) and Davies--Bouldin (DB, $\downarrow$) scores}
\label{tab:cluster_metrics}
\small
\setlength{\tabcolsep}{4pt}
\begin{tabular}{lcccc}
\toprule
\multirow{2}{*}{Benchmark} &
\multicolumn{2}{c}{Silhouette $\uparrow$} &
\multicolumn{2}{c}{DB $\downarrow$} \\
\cmidrule(lr){2-3}
\cmidrule(lr){4-5}
& \textit{Spectra} & RGB & \textit{Spectra} & RGB \\
\midrule
CDF$_{\mathrm{v1}}$ & \textbf{0.1337} & 0.0561 & \textbf{2.517} & 3.939 \\
CDF$_{\mathrm{v2}}$ & \textbf{0.1155} & 0.0421 & \textbf{2.632} & 4.620 \\
DFD               & \textbf{0.2306} & 0.0808 & \textbf{1.690} & 3.178 \\
FaceShifter       & \textbf{0.1059} & 0.0330 & \textbf{2.741} & 5.153 \\
\midrule
Mean              & \textbf{0.1465} & 0.0530 & \textbf{2.395} & 4.222 \\
\bottomrule
\end{tabular}
\end{table}

\subsection{Can \textit{\textit{Spectra}} be Extended to Presentation Attack Detection?}

\begin{figure}[t]
    \centering
    \includegraphics[width=0.9\columnwidth]{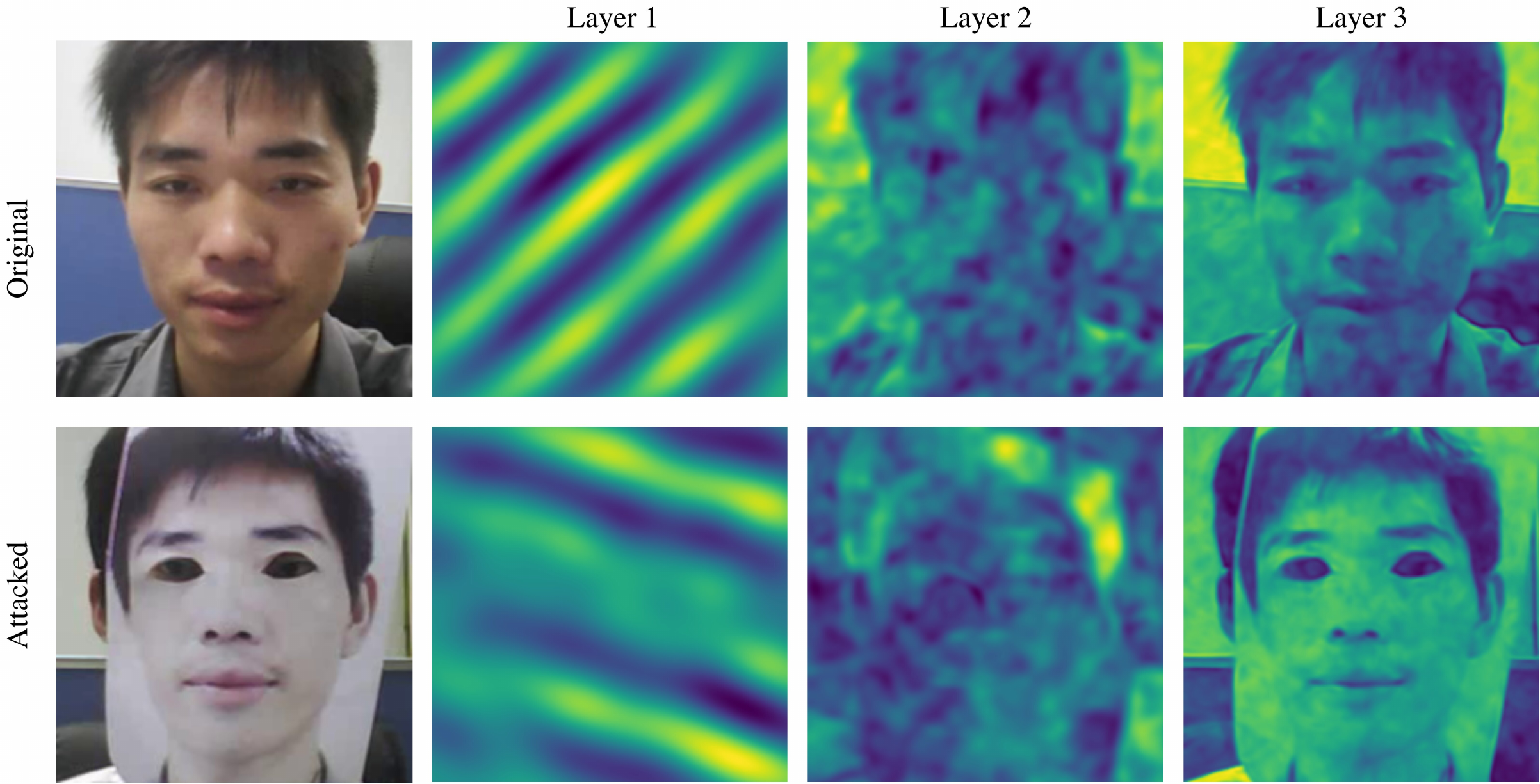}
    \captionsetup{font=small}
\vskip-5pt    \caption{\textbf{INR layer-wise decomposition.} Genuine samples produce smooth activations across layers, while attacks yield noisier intermediate responses and less consistent structure. }
    \label{fig:inr_decomp_real_attack}
\end{figure}

Deepfake manipulation and presentation attacks arise from fundamentally different processes. Deepfakes introduce localized, semantic-level inconsistencies,  a manipulated region embedded within an otherwise natural image. Presentation attacks instead arise from low-level imaging physics, recapture noise, printing artifacts, and shading or reflectance mismatches introduced when a face is presented to a sensor rather than captured directly. {\textit{Spectra}} learns features by detecting deviations from the natural spectral and structural properties of images, rather than relying on artifacts specific to deepfakes.
This raises a natural question, does the same representation transfer to presentation attack detection, a setting governed by entirely different forgery mechanics?   \Cref{fig:inr_decomp_real_attack} visualizes the layer-wise INR decomposition for a genuine (top) and a presentation-attacked (bottom) sample. Even at Layer~1, the samples exhibit distinct low-level, coordinate-dependent sinusoidal responses. The two diverge largely by Layer~2, the attacked sample develops a sharply localized bright region near the forehead absent in the genuine sample, consistent with a non-ideal reflectance pattern introduced by the recapture process. By Layer~3, the genuine sample maintains coherent, evenly distributed facial structure, while the attacked sample's reconstruction is comparatively less uniform, a subtler distinction than the artifact patterns observed for deepfakes, reflecting that presentation attacks perturb image statistics more globally and less locally than semantic manipulation does. We evaluate this transfer on four widely used face anti-spoofing benchmarks: CASIA-MFSD (C)~\cite{zhang2012face}, Idiap Replay-Attack (I)~\cite{chingovska2012effectiveness}, MSU-MFSD (M)~\cite{wen2015face}, and OULU-NPU (O)~\cite{boulkenafet2017oulu}, under the standard leave-one-domain-out protocol, each dataset in turn serves as the held-out target while the model trains on the remaining three, testing generalization across sensors, capture conditions, and attack types. As shown in \Cref{tab:pad_results}, \textit{\textit{Spectra}} achieves the best average performance (97.17) across all four protocols, and the best result on three of the four (M, C, O), trailing only on I, where two prior methods report marginally higher scores. This indicates that the reconstruction-aware representation generalizes beyond deepfake detection to a forgery setting with substantially different underlying physics, without any task-specific architectural modification.

\begin{table}[t]
\centering
\caption{\textbf{Cross-dataset PAD performance (\% AUC).} M, C, I,
and O denote the four evaluation protocols. Best and second-best
results are highlighted in \textbf{bold} and \underline{underlined},
respectively.}
\label{tab:pad_results}
\small
\setlength{\tabcolsep}{5pt}
\begin{tabular}{lccccc}
\toprule
\textbf{Method} & \textbf{M} & \textbf{C} & \textbf{I} & \textbf{O} & \textbf{Avg} \\
\midrule

RP'21~\cite{yu2021revisiting}
    & 87.43
    & 86.76
    & 86.38
    & 92.43 & 88.25 \\

FG'22~\cite{shao2022federated}
    & 91.25
    & 80.58
    & \textbf{95.34}
    & 89.85 & 89.25\\

FDGAL'24~\cite{huang2023face}
    & 93.29
    & 88.10
    & 91.92
    & 90.54 & 90.96\\

MRGA'23~\cite{zhu2023multiscale}
    & 89.83
    & 88.36
    & \underline{94.02}
    & 90.86 & 90.77 \\

ST'24~\cite{li2024face}
    & 90.00
    & 89.98
    & 87.81
    & 92.73 & 90.13\\

HPDR'24~\cite{hu2024rethinking}
    & 96.02
    & 94.42
    & 92.49
    & \underline{95.26} & 94.55\\

CA-FAS'25~\cite{long2025confidence}
    & \underline{97.42}
    & \underline{94.55}
    & 93.67
    & \underline{94.53} & 95.04\\

\rowcolor{green!10}\textbf{\textit{Spectra}}
    & \textbf{98.85}
    & \textbf{99.84}
    & 93.42
    & \textbf{96.58} & \textbf{97.17} \\
\bottomrule
\end{tabular}
\end{table}

\section{Additional Experiments and Ablations}

The supplementary material provides additional details that complement the main paper. This includes an expanded related work section, a detailed description of the INR architecture, and the rationale behind key design choices such as activation functions, network depth, hidden-layer dimensionality, and the proposed hypernetwork-based acceleration mechanism. We further provide complete implementation details, including network hyperparameters and descriptions of the major architectural components. Extensive ablation studies are conducted to analyze the contributions of the CLIP backbone, reconstruction-aware heatmaps, multi-level decomposition, multi-scale semantic representations, and the employed loss functions. Finally, additional qualitative visualizations are presented to illustrate the ability of the proposed INR decomposition to capture hierarchical structural information across multiple representation levels for both deepfake detection and face presentation attack detection.

\vspace{-5pt}

\section{Conclusion}
We introduced \textit{Spectra}, a reconstruction-aware forensic framework that combines implicit neural representation dynamics with CLIP semantics to detect synthetic images. By jointly modeling \emph{what} an image depicts and \emph{how} it is reconstructed, \textit{Spectra} learns representations that are more robust than appearance-based features alone. Extensive experiments demonstrate state-of-the-art performance on cross-dataset deepfake detection and strong generalization to presentation attack detection without any task-specific redesign. Furthermore, the learned embeddings exhibit substantially improved class separability, indicating that reconstruction-aware features capture fundamental signatures of synthetic content that transfer across different manipulation and attack types.

\clearpage

{
    \small
    \bibliographystyle{ieeenat_fullname}
    \bibliography{main}
}

\end{document}